# Design of Sensor Fusion Driver Assistance System for Active Pedestrian Safety

# I-Hsi Kao[1, *], Ya-Zhu Yian[1], Jian-An Su[1], Yi-Horng Lai[1], Jau-Woei Perng[1], Tung-Li Hsieh[2], Yi-Shueh Tsai[3],and Min-Shiu Hsieh[3]

[1] National Sun Yat-sen University, Kaohsiung, Taiwan
[2] Wenzao Ursuline University of Languages, Kaohsiung, Taiwan
[3] Automotive Research and Testing Center (ARTC), Changhua, Taiwan
*Corresponding author: flush0129@gmail.com

**Abstract:** In this paper, we present a parallel architecture for a sensor fusion detection system that combines a camera and 1D light detection and ranging (lidar) sensor for object detection. The system contains two object detection methods, one based on an optical flow, and the other using lidar. The two sensors can effectively complement the defects of the other. The accurate longitudinal accuracy of the object's location and its lateral movement information can be achieved simultaneously. Using a spatio-temporal alignment and a policy of sensor fusion, we completed the development of a fusion detection system with high reliability at distances of up to 20 m. Test results show that the proposed system achieves a high level of accuracy for pedestrian or object detection in front of a vehicle, and has high robustness to special environments.

## 1. Introduction

Every year, more than 1.2 million people around the world die in traffic accidents, and among these casualties, 22% are pedestrians. In 2013, 4,735 pedestrians were killed and 66,000 seriously injured in traffic accidents in the United States alone. For these reasons, counterplans are needed to reduce pedestrian accidents. Many types of automotive electronic systems, such as adaptive cruise control, autonomous emergency braking systems, and automatic parking systems, have been widely used in active or passive pedestrian safety systems. Numerous vehicles have been installed with autonomous emergency braking systems as a protective mechanism. This type of system helps in reducing the number of accidents by alerting the driver and controlling the automatic brake actuator before an accident occurs. Moreover, since 2016, the European New Car Assessment Program (NCAP) has stipulated the use of an autonomous emergency braking (AEB) pedestrian system as a standard requirement [1].

Using simulation testing, J. Lenard predicted and verified the reliability of an AEB system under various environments [2]. The results showed that an AEB system is highly valuable in reducing traffic accidents. H. Hamdane highlighted the issue of active pedestrian safety through a reconstruction of 100 real accident cases involving a vehicle and pedestrian. The functionality of an AEB system was analyzed according to the reconstructed crash cases. Based on the use of established graphs, it is possible to evaluate the efficiency of an active safety system and define its specifications [3].

To develop a pedestrian AEB system, pedestrian detection technology is critical. A pedestrian protection system can be developed using sensing technology such as radar, cameras, or light detection and ranging (lidar) [4]-[8]. However, each sensor has its own performance limitations, such as the detection range, accuracy, and pedestrian status for automotive applications. Such performance limitations also limit the system performance. For these reasons, a fusion of such sensors needs to be applied.

P. M. Hsu presented a sensor fusion method focusing on a collision avoidance system that is capable of highly accurate object detection by blending a camera and 2D lidar sensor with the aid of a pixel analysis. This fusion system is aimed at locating stationary or moving objects, such as in front of a car [9]. In [10], L. Huang proposed a novel vehicle detection system based on a tight integration of lidar and camera images. 2D lidar provides excellent range information on different objects. On the other hand, computer vision imagery allows for better recognition, but does not provide a high-resolution range. This type of system is quite useful for the modeling and prediction of traffic conditions over a variety of roadways. C. Premebida presented a sensorial-cooperative architecture to detect, track, and classify entities in semi-structured outdoor scenarios for intelligent vehicles. The detection and tracking phases are applied in the laser (using a Gaussian mixture model classifier) and vision (using an AdaBoost classifier) spaces [11]. F. Garcia, D.

Martin, A. de la Escalera, and J. M. Armingol presented a novel sensor fusion system that provides intelligent vehicles with augmented environment information and knowledge, enabled by a vision-based system, laser sensor, and a global positioning system. The main goal of the methodology for intelligent vehicles is to overcome the limitations of each sensor, providing a robust and reliable safety application for a road environment [12]. The above examples show the advantages of a sensor fusion system.

In recent years, an AEB system for pedestrian detection has been frequently discussed by researchers [1], [13], [14]. In car accidents involving pedestrians, the mortality rate of the pedestrians is much higher than that of the people inside the vehicle. For this reason, the detection of pedestrians is more important than the detection of vehicles in an AEB system.

A lidar sensor has been widely used in AEB systems to measure the distance to surrounding objects. Among the different types of lidar devices, the SRL-1 short-range lidar sensor from Continental (Germany) has the advantage of cost efficiency. Unlike expensive 2D or 3D lidar systems installed with a complicated reflector mechanism, the SRL-1 uses three independent infrared laser beams to measure the distance to an object. The SRL-1 has been successfully selected as a standard application by many different automobile companies. However, considering the limitations of the measurement capability of the SRL-1 (up to 13.5 m at an angle of 27°) [15], the additional use of a sensor fusion methodology can extend its detection performance.

This study presents a sensor fusion system for AEB detection. To improve the accuracy and detection distance, the system combines an SRL-1 lidar sensor and a camera. The remainder of this paper is organized as follows. Section 2 details the configuration of the sensor fusion system and spatio-temporal alignment between the two sensors. In Section 3, the methodology of detection for the lidar and camera sensors and the policy of sensor fusion system are presented. The experiment results of the proposed system are described in Section 4. Finally, Section 5 provides some concluding remarks regarding this study.

## 2. System Structure

A parallel architecture of sensor fusion was devised for the proposed system, in which there is no predefined main sensor used. Under different circumstances, the system should choose the preferable sensor to apply. The implementation of the system planning is shown in Fig. 1.

Because an SRL-1 is a short-range lidar sensor, it has a poor response when the object is at a distance of much greater than 10 m. However, the SRL-1 lidar sensor has a higher degree of stability in special environments such as darkness. On the other hand, the camera has farther and wider horizons than the SRL-1 lidar sensor. By fusing the two sensors, we complete a fusion detection system with high reliability at a distance of up to 20 m. The detection algorithm selects the closest in frontal object (CIFO) spatio-temporal alignment.

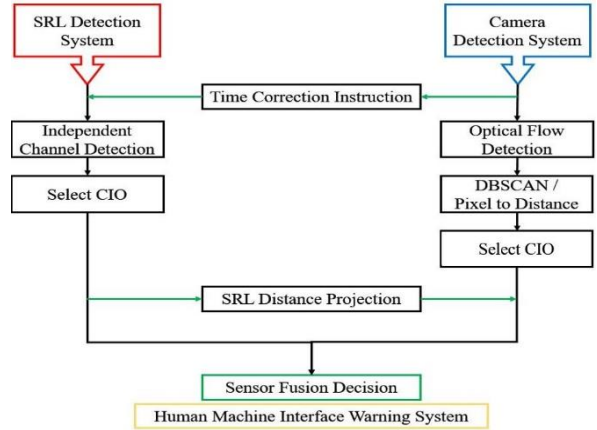

Fig.1. Fusion system

### A. Camera

A Logitech HD PRO Webcam C920R camera was used in this study, as shown in Fig. 2. The specifications of the C920R are listed in Table. 1. Using an optical flow method, a maximum of five targets can be detected.

Table 1. Camera Specifications

| Parameter | Specification |
|---|---|
| Resolution | $640 \times 480$ |
| Diagonal Total Field of View (FOV) | $78^o$ |
| Frame Rate | 30 fps |
| Detection Range | 20 m |

### B. SRL-1 lidar sensor

An SRL-1 lidar sensor is used in this study, as shown in Fig. 3. The SRL-1 avoids the contradiction between excellent measurement performance and high degree of operational safety, and is able to determine the distance to an object with a very high repetition rate while, depending on the driving speed, avoiding the risk of collision. An SRL-1 is equipped with self-monitoring and cyclic realized self-diagnosis capabilities. Possible hazards are recognized by the sensor and displayed automatically. The specifications of the SRL-1 are listed in Table. 2.

Table 2. Lidar Specifications

| Parameter | Specification |
|---|---|
| Distance Range | 1.0 - 10.0 m ($\pm 0.1$) |
| Total Field of View (FOV) | $27^o(H) \times 11^o(V)$ |
| Repetition Rate | 100 Hz |

### C. Spatio-temporal alignment

Using a sensor fusion algorithm, both a highly accurate detection of an object longitudinally and the lateral movement information can be achieved simultaneously. The proposed sensor fusion algorithm consists of two functions: spatio-temporal alignment and the policy of sensor fusion.

Similar to [1], we adopt a multi-model approach to achieve good spatial alignment precision. First, the distance of the reference object is divided into 1 m steps between 2 to 10 m away and recorded on a horizontal plane. Second, the position of the reference object on the image captured by a camera can be regarded as $R_c$. The detection range of the SRL-1 can be regarded as $R_l$. A spatial alignment method is proposed to estimate the transformation between the camera distance and lidar sensor distance through a curve fitting.

Temporal alignment depends on the processing time of the camera. The frame rate of the camera is about 30 fps. When each frame calculation is finished, an externally temporal calibration trigger command will be sent to the SRL-1 lidar sensor for synchronous detection with the camera.

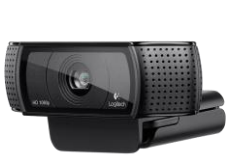

Fig. 2. Logitech C920R

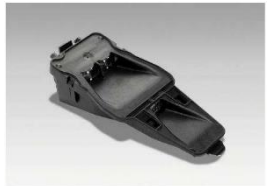

Fig. 3. Continental SRL

## 3. Detection Algorithm

### A. Detection algorithm of SRL-1 Lidar

Although there no horizontal or vertical information is provided by the SRL-1 lidar sensor, the distance and velocity of the object can be detected using three independent measuring channels. To reduce the false alarm rate, a detecting management algorithm is proposed to calculate the standard deviation of distance in each independent channel with 30 continuous frames. According to its data sheet, the standard deviation of the SRL-1 lidar sensor distance is 10 cm. In each channel, any measurement beyond twice the criterion will be judged as a false alarm and eliminated.

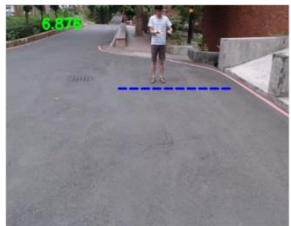

Fig. 4. Detection result of SRL-1 lidar sensor

The SRL-1 lidar sensor can detect a maximum of three tracking objects. The detecting management algorithm selects the closest in object (CIO) from three independent channel outputs. The projective transformation of distance for CIO will then be displayed in the image plane of the camera (as shown in Fig. 4).

### B. Detection algorithm for camera

Detection by a camera contains the region of interest (ROI), corners, and optical flow. Through such a method, an object within the frame is detectable. After we recognize the pixels of the object within the frame, we transfer the bottom pixel into the distance using a curve fitting function. Finally, we obtain the distance of the object through a vision signal only.

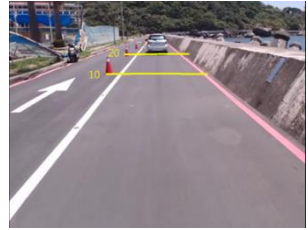 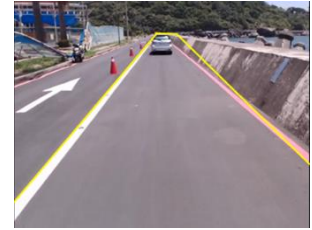

Fig. 5. Farthest Pixel Searching       Fig. 6. ROI

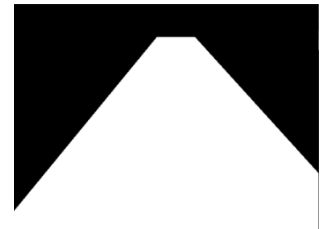

Fig. 7. ROI with thresholding

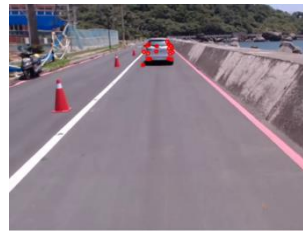 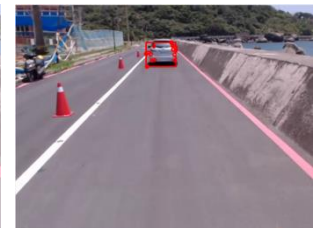

Fig. 8. Corner detection       Fig. 9. Optical flow and DBSCAN

In this study, we obtain the ROI by calibrating the area of interest in the image. First, by placing the object 30 m in front of the vehicle, as shown in Fig. 5, we can recognize the farthest pixel of concern. However, the calibration area only contains the lane up to 30 m in front of the vehicle. According to the description, we can obtain four point coordinates to draw the ROI, as indicated in Fig. 6.

After obtaining the ROI, we create an image mask, which is the same scope as the ROI, and process the image using thresholding, as indicated in Fig. 7. Using this method, an object that is out of the ROI will be disregarded by the system.

To complete the corner detection, we set the number of corner points, as well as the corner detection function and its parameters. The detection results are shown in Fig. 8. The detection function is completed using the Shi-Tomasi algorithm [16].

The main concept of an optical flow is a change in the corner features of the previous and following frames, and the movement and tracking of the pixel position.

In [17], an optical flow was applied to cell tracking in the biomedical field. In [18], Wang calculated the motion equation using an optical flow method, tracking the feature points, and applied the least squares method to compensate the visual jitter generated by a robot after a collision. This equation shows good results in detecting objects using an optical flow.

In this study, after detecting the corners in the ROI, DBSCAN is applied to cluster the points as necessary. The corners will be classified after DBSCAN is used. The classification results show that the object has been detected, as indicated in Fig. 9.

To obtain the distance to the objects, we should calculate a pixel that is at the bottom of the DBSCAN result block. We collected the distance results of the object corresponding to the point on the $y$ pixel for 53 data sets and curved the function through curve fitting. The curve fitting function is shown in (1):

$$x = ay^8 + by^7 + cy^6 + dy^5 + ey^4 + fy^3 + gy^2 + hy + i \qquad (1)$$

where y is the pixel of the bottom object box, and x is the distance to the object. By calculating the parameters in (1), we get a function that can obtain the distance when inputting the y pixel.

The camera can detect a maximum of five targets. According to the statistics from our experimental dataset, the standard deviation in the detection of the proposed image processing algorithm is 20 cm with 30 continuous frames. Any measurement beyond twice the criterion will be judged as a false alarm and therefore eliminated. Similar to the detection management algorithm of lidar, the CIO in the camera and SRL-1 will be fused with the sensor fusion policy.

### C. Sensor Fusion Policy

The tracking data of the CIO from the SRL-1 lidar sensor and camera are fused based on a sensor fusion policy. The most important function within the sensor fusion algorithm is the main signal switch, which relies on the signal from the camera or SRL-1 lidar sensor. According to its specifications, we realize that the distance range of the SRL-1 lidar sensor is insufficient. For this reason, the camera is used as the main sensor when detecting an object located between 9 to 20 m away.

However, the detection results of the SRL-1 lidar sensor are more accurate than those of the camera when the object is between 1 to 9 m. In this case, the SRL-1 lidar sensor becomes the main sensor of the system. In certain cases, the SRL-1 lidar sensor also becomes the main sensor. For example, when the vehicle is in a dark environment, the camera does not have clear vision capability for object detection, and the SRL-1 lidar sensor will replace the responsibility of the camera sensor.

The fusion algorithm is shown in Fig. 10, where L[x] indicates a subset of the object distance detected by the SRL-1 lidar sensor, and C[y] is a subset of the object distance detected by the camera. When there is no detection signal from the SRL-1 lidar sensor, the camera becomes the main sensor. On the other hand, when the camera obtains no detection results, the lidar sensor becomes the main sensor. If neither sensor achieves a detection result, we apply the slightest detection result so as to maintain a safe situation.

The human-machine interface of the proposed system includes the detected distance from the SRL-1 lidar sensor (green), the distance computed by the camera (red), and two types of warning conditions. A CIO distance beyond 10 m will be attributed as the first warning condition. A visible yellow warning is activated and displayed on the human machine interface (HMI). A CIO distance of less than 10 m will be attributed as the second warning condition. A visible red warning is then activated and displayed on the HMI.

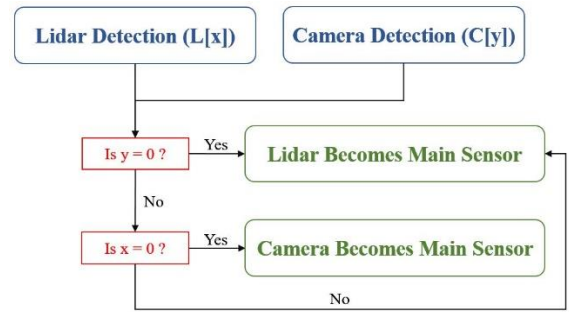

Fig. 10. Fusion algorithm

## 4. Experiment

In our experiment, the camera and SRL-1 lidar sensor are placed on a golf cart and used to detect pedestrians in front of an object, as shown in Fig. 11.

The test scenario focused on pedestrian detection. Following the standard procedure of the Euro NCAP, the conditions of the test track were dry with no visible moisture on the surface, at an ambient temperature of above 20°C and below 30°C.

To validate the evaluated algorithm, we set the vehicle speed to 15 km/h. The following Euro NCAP VRU test scenarios were carried out:

Car-to-VRU Farside Adult (CVFA) – A collision in which a vehicle moves forward toward an adult pedestrian crossing the vehicle's path while running from the farside, and where the front structure of the vehicle strikes the pedestrian at 50% of the vehicle width when no braking action is applied.

Car-to-VRU Nearside Adult (CVNA-25) – A collision in which a vehicle travels forward toward an adult pedestrian crossing the vehicle's path while walking from the nearside, and where the front structure of the vehicle strikes the pedestrian at 25% of the vehicle width when no braking action is applied.

Car-to-VRU Nearside Adult (CVNA-75) A collision in which a vehicle travels forward toward an adult pedestrian crossing the vehicles' path while walking from the nearside, and where the front structure of the vehicle strikes the pedestrian at 75% of the vehicle width when no braking action is applied.

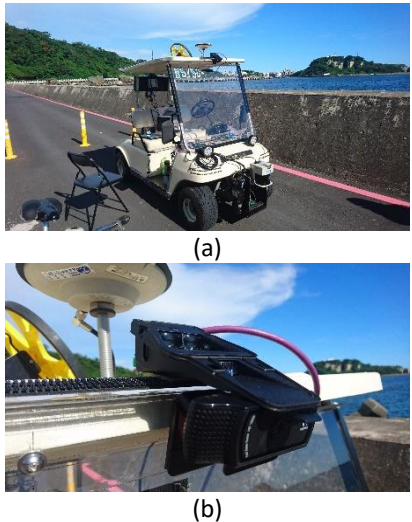

(a)

(b)

Fig. 11. Experiment device: (a) experiment vehicle and (b) experiment sensors.

The results of the Euro NCAP VRU test scenarios are shown in Fig. 12. A CIO distance of beyond 10 m will be attributed as the first level warning condition. A visible yellow warning is activated and displayed on the HMI. On the other hand, a CIO distance of less than 10 m will be attributed as the second level warning condition. A visible red warning is activated and displayed on the HMI.

To evaluate the system's robustness, a more precise evaluation was conducted under various conditions on a real road. There were six special cases chosen in our experiment: at night, in the presence of shadows, with a black lighting, more than one pedestrian present, and a pedestrian sitting on a chair. The results of these special cases are shown in Fig. 13.

In Fig. 13, the number on the left is the distance detected by the SRL-1, and the number on the right is the distance detected by the camera. The system can detect pedestrians at night, as shown in Fig. 13(a). When the pedestrians are in the presence of a shadow, they can also be detected, as shown in Fig. 13(b). In black-lighted environments, the camera often determines an incorrect detection distance, as shown in Fig. 13(c). In this case, the SRL-1 becomes the main sensor through the fusion system. In Fig. 13(d), although a number of pedestrians are present, the camera sensor detects all of them but only shows the nearest distance. The SRL-1 detects the nearest pedestrian when more than one pedestrian is present.

As shown in Fig. 13(e), although some detection systems are unable to detect pedestrians sitting on a chair [1], our proposed system is able to achieve this type of detection.

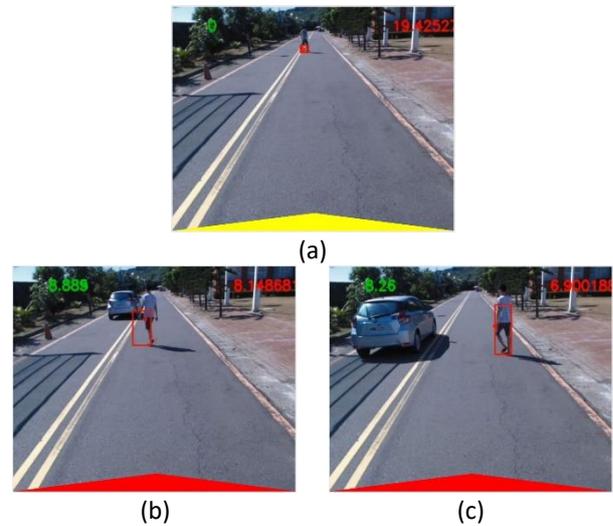

Fig. 12. Test protocol of the Euro NCAP VRU: (a) CVFA scenario where an adult runs from the farside, (b) CVNA-75, where an adult walks from the nearside, and (c) CVNA-25, where an adult walks from the nearside.

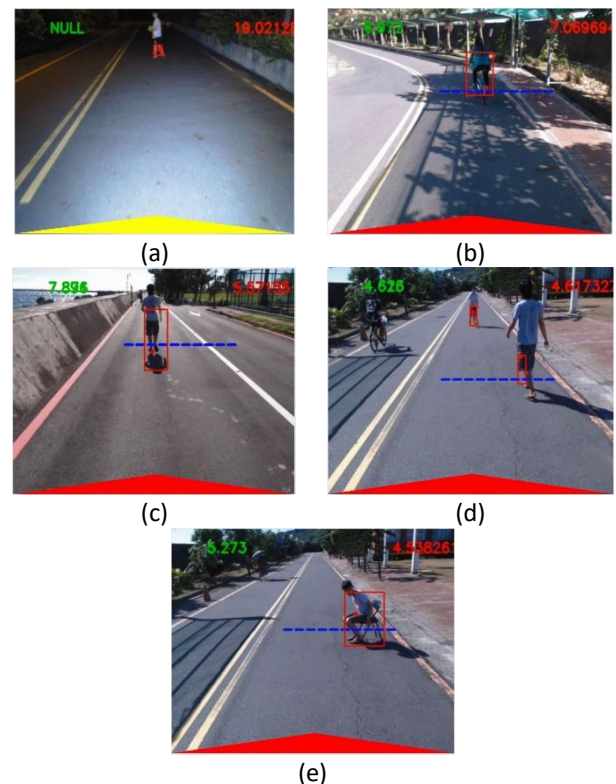

Fig. 13. Fusion detection results: (a) at night, (b) in the presence of shadows with a bike, (c) with a black lighting, (s) more than one pedestrian present, and (e) pedestrian sitting on a chair.

## 5. Conclusion

In the future, vehicles will be expected to have a variety of safety functions for autonomous driving. A pedestrian AEB system is one if such functions. The study presented in this paper aims at the detection of pedestrians using a fusion of an SRL-1 sensor and a camera.

When these two sensors are used to detect objects independently, their detection results are impractical. The SRL-1 can only detect objects within a distance of 10 m, and the camera can only detect objects within a distance of 20 m, and shows poor results for objects within a distance of 10 m. For these reasons, the fusion of these two types of sensors may obtain good results. Although our idea was proved experimentally, there are still certain cases that cause a system failure. For instance, the algorithm used by the camera sensor is insufficiently stable to adapt to all types of environments.

Our future work will improve on the limitations of the proposed sensor fusion concept. For example, by adding Bayes' theorem to determine the results of the optical flow, the number of false positives may be decreased.

## 6. Acknowledgement

Research supported by the Ministry of Science and Technology (no. MOST 105-2218-E-110-009).

## 7. References

[1] F. Garcia, D. Martin, A. de la Escalera, and J. M. Armingol, "Sensor fusion methodology for vehicle detection," *IEEE Intelligent Transportation System Magazine*, vol. 9, 2017, pp. 123-133.

[2] J. Lenard, A. Badea-Romero, and R. Danton, "Typical pedestrian accident scenarios for the development of autonomous emergency brake test protocols," *Accident Analysis and Prevention*, 2014, pp. 73-80.

[3] H. Hamdane, T. Serre, C. Masson, and R. Anderson, "Issues and challenges for pedestrian active safety systems based on real world accidents," *Accident Analysis and Prevention*, 2015, pp. 53-60.

[4] K. C. Fuerstenberg, K. C. J. Dietmayer, and V. Willhoeft, "Pedestrian recognition in urban traffic using a vehicle based multilayer laserscanner," *IEEE Intelligent Vehicle Symposium*, 2002, pp. 31-35.

[5] T. Gandhi and M. M. Trivedi, "Pedestrian protection systems: Issues, survey, and challenges," *IEEE Transactions on Intelligent Transportation Systems*, vol. 8, 2007, pp. 413-430.

[6] G. Samuel, C. Paul, B. Christophe, C. Thierry, and T. Laurent, "Pedestrian detection and tracking in an urban environment using a multilayer laser scanner," *IEEE Transactions on Intelligent Transportation System*, vol. 11, 2010, pp. 579-588.

[7] D. F. Llorca et al., "Autonomous pedestrian collision avoidance using a fuzzy steering controller," *IEEE Transactions on Intelligent Transportation Systems*, vol. 12, 2011, pp. 390-401.

[8] A. Mukhtar, L. Xia, and T. B. Tang, "Vehicle detection technique for collision avoidance systems: A review," *EEE Transactions on Intelligent Transportation System*, vol. 16, 2015, pp. 2318-2338.

[9] P. M. Hsu, M. H. Li, and Y. F. Su, "Object detection and recognition by using sensor fusion," *IEEE International Conference on Control & Automation*, 2014, pp. 56-60.

[10] L. Huang, and M. Barth, "Tightly-coupled lidar and computer vision integration for vehicle detection," *IEEE Intelligent Vehicles Symposium*, 2009, pp. 604–609.

[11] C. Premebida, G. Monteiro, U. Nunes, and P. Peixoto, "A lidar and vision-based approach for pedestrian and vehicle detection and tracking" *IEEE Intelligent Transportation Systems Conference*, 2007, pp. 1044-1049.

[12] M. K. Park, S. Y. Lee, C. K. Kwon, and S. W. Kim, "Design of pedestrian target selection with funnel map for pedestrian AEB system," *IEEE Transactions on Vehicular Technology*, vol. 66, 2017, pp. 3597-3609.

[13] T. Y. Han, B. C. Song, "Night vision pedestrian detection based on adaptive preprocessing using near infrared camera," *IEEE International Conference on Electronic-Asia*, 2016, pp. 1-3.

[14] A. López Rosado, S. Chien, L. Li, Q. Yi, Y. Chen, and R. Sherony, "Certain and critical speed for decision making in tests of pedestrian automatic emergency braking transportation systems," *IEEE Transactions on Intelligent Transportation Systems*, vol. 18, 2017, pp. 1358-1370.

[15] http://www.conti-online.com/www/industrial_sensors_de_en/themes/srl_1_en.html

[16] J. Shi and C. Tomasi, "Good features to track," *IEEE Computer Vision and Pattern Recognition, Proceedings CVPR '94. Computer Society Conference*, 1994, pp. 593-600.

[17] D. Guo, AL. van de Ven, X. Zhou, "Red Blood Cell Tracking Using Optical Flow Methods," *IEEE Journal of Biomedical and Health Informatic,* vol. 18, 2014, pp. 991-998.

[18] B. Wang, Y. Xu, X. Chen, Y. Jin, "Collision jitter compensation for mobile robots vision based on optical flow estimation," *IEEE International Conference on Cyber Technology in Automation, Control and Intelligent Systems*, 2015, pp. 492-496.